\documentclass[11pt,a4paper]{article}
\usepackage[hyperref]{acl2020}
\usepackage{times}
\usepackage{enumitem}
\usepackage{latexsym}
\usepackage{graphicx}
\usepackage{amssymb}
\usepackage[ruled,vlined,linesnumbered]{algorithm2e}

\usepackage{amsmath}
\usepackage{mathtools}
\DeclareMathOperator*{\argmax}{arg\,max}

\usepackage[noend]{algpseudocode}
\usepackage{microtype}

\aclfinalcopy 

\newcommand{\dataset}[1]{\textsc{#1}}
\newcommand{\pixelhelp}{\dataset{PixelHelp}}
\newcommand{\howto}{\dataset{AndroidHowTo}}
\newcommand{\ricosynth}{\dataset{RicoSCA}}

\title{Mapping Natural Language Instructions to Mobile UI Action Sequences}

\author{Yang Li\space\space\space\space\space Jiacong He\space\space\space\space\space Xin Zhou\space\space\space\space\space Yuan Zhang\space\space\space\space\space Jason Baldridge\\
  Google Research, Mountain View, CA, 94043 \\
  \texttt{\{liyang,zhouxin,zhangyua,jasonbaldridge\}@google.com}}

\date{}

\begin{document}
\maketitle
\begin{abstract}
We present a new problem: grounding natural language instructions to mobile user interface actions, and create three new datasets for it. For full task evaluation, we create \pixelhelp, a corpus that pairs English instructions with actions performed by people on a mobile UI emulator. To scale training, we decouple the language and action data by (a) annotating action phrase spans in HowTo instructions and (b) synthesizing grounded descriptions of actions for mobile user interfaces. We use a Transformer to extract action phrase tuples from long-range natural language instructions. A grounding Transformer then contextually represents UI objects using both their content and screen position and connects them to object descriptions. Given a starting screen and instruction, our model achieves 70.59\% accuracy on predicting \textit{complete} ground-truth action sequences in \pixelhelp.  
\end{abstract}

\begin{figure}[t]
  \centering
  \includegraphics[width=1.0\linewidth]{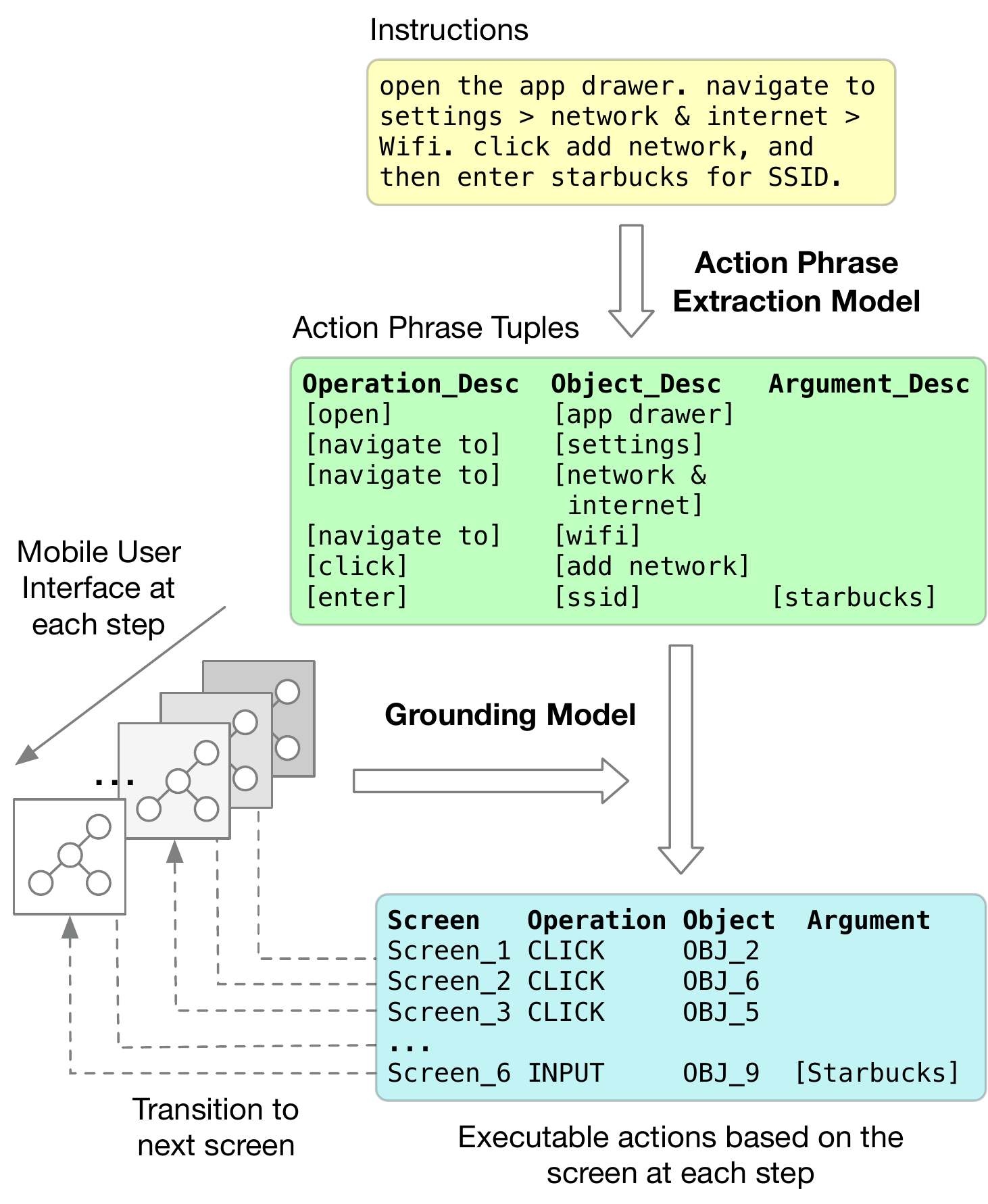}
  \caption{Our model extracts the phrase tuple that describe each action, including its operation, object and additional arguments, and grounds these tuples as executable action sequences in the UI.}
  \label{fig:workflow}
\end{figure}

\section{Introduction}
\label{sec:introduction}

Language helps us work together to get things done. People instruct one another to coordinate joint efforts and accomplish tasks involving complex sequences of actions. This takes advantage of the abilities of different members of a speech community, e.g. a child asking a parent for a cup she cannot reach, or a visually impaired individual asking for assistance from a friend. 
Building computational agents able to help in such interactions is an important goal that requires true language grounding in environments where action matters.

An important area of language grounding involves tasks like completion of multi-step actions in a graphical user interface conditioned on language instructions \citep{Branavan:2009:RLM:1687878.1687892,branavan-etal-2010-reading,liu2018workflow,gur2018learning}. These domains matter for accessibility, where language interfaces could help visually impaired individuals perform tasks with interfaces that are predicated on sight. This also matters for \textit{situational impairment} \citep{sarsenbayeva:2018} when one cannot access a device easily while encumbered by other factors, such as cooking.

We focus on a new domain of task automation in which natural language instructions must be interpreted as a sequence of actions on a mobile touchscreen UI. Existing web search is quite capable of retrieving multi-step natural language instructions for user queries, such as ``How to turn on flight mode on Android.'' Crucially, the missing piece for fulfilling the task automatically is to map the returned instruction to a sequence of actions that can be automatically executed on the device with little user intervention; this our goal in this paper. This task automation scenario does not require a user to maneuver through UI details, which is useful for average users and is especially valuable for visually or situationally impaired users. The ability to execute an instruction can also be useful for other scenarios such as automatically examining the quality of an instruction.

Our approach (Figure \ref{fig:workflow}) decomposes the problem into an \textit{action phrase-extraction} step and a \textit{grounding} step. The former extracts operation, object and argument descriptions from multi-step instructions; for this, we use Transformers \citep{DBLP:journals/corr/VaswaniSPUJGKP17} and test three span representations. The latter matches extracted operation and object descriptions with a UI object on a screen; for this, we use a Transformer that contextually represents UI objects
and grounds object descriptions to them.

We construct three new datasets
\footnote{Our data pipeline is available at \url{https://github.com/google-research/google-research/tree/master/seq2act}.}.
To assess full task performance on \textit{naturally occurring} instructions, we create a dataset of 187 multi-step English instructions for operating Pixel Phones and produce their corresponding action-screen sequences using annotators. For action phrase extraction training and evaluation, we obtain English How-To instructions from the web and annotate action description spans. A Transformer with spans represented by sum pooling \citep{pmlr-v97-li19e} obtains 85.56\% accuracy for predicting span sequences that completely match the ground truth. To train the grounding model, we synthetically generate 295k single-step commands to UI actions, covering 178K different UI objects across 25K mobile UI screens.

Our phrase extractor and grounding model together obtain 89.21\% partial and 70.59\% complete accuracy for matching ground-truth action sequences on this challenging task. We also evaluate alternative methods and representations of objects and spans and present qualitative analyses to provide insights into the problem and models.

\section{Problem Formulation}
\label{sec:formulation}

Given an instruction of a multi-step task, $I=t_{1:n}=(t_{1},t_{2},...,t_{n})$, where $t_i$ is the $i$th token in instruction $I$, we want to generate a sequence of automatically executable actions, $a_{1:m}$, over a sequence of user interface screens $S$, with initial screen $s_1$ and screen transition function $s_{j}{=}\tau(a_{j-1}, s_{j-1})$: 

\begin{equation}
    \label{eq:inst2acts}
    p(a_{1:m}|s_{1},\tau, t_{1:n})= \prod_{j=1}^{m}p(a_{j}|a_{<j},s_{1},\tau,t_{1:n})
\end{equation}

An action $a_j=[r_j,o_j,u_j]$ consists of an operation $r_j$ (e.g. \texttt{Tap} or \texttt{Text}), the UI object $o_j$ that $r_j$ is performed on (e.g., a button or an icon), and an additional argument $u_j$ needed for $o_j$ (e.g. the message entered in the chat box for \texttt{Text} or \texttt{null} for operations such as \texttt{Tap}).
Starting from $s_1$, executing a sequence of actions $a_{<j}$ arrives at screen $s_j$ that represents the screen at the $j$th step: $s_{j}=\tau(a_{j-1}, \tau(...\tau(a_1, s_1)))$:

\begin{equation}
    \label{eq:inst2acts2}
    p(a_{1:m}|s_{1},\tau, t_{1:n})= \prod_{j=1}^{m}p(a_{j}|s_{j},t_{1:n})
\end{equation}

Each screen $s_j=[c_{j,1:|s_j|}, \lambda_j]$ contains a set of UI objects and their structural relationships. 
$c_{j,1:|s_j|}=\{c_{j,k}\mid 1\leq{k}\leq{|s_j|\}}$, where $|s_j|$ is the number of objects in $s_j$, from which $o_j$ is chosen. $\lambda_j$ defines the structural relationship between the objects. This is often a tree structure such as the \textit{View} hierarchy for an Android interface\footnote{\url{https://developer.android.com/reference/android/view/View.html}} (similar to a DOM tree for web pages).

An instruction $I$ describes (possibly multiple) actions. Let $\bar{a}_{j}$ denote the phrases in $I$ that describes action $a_j$. $\bar{a}_{j}=[\bar{r}_{j}, \bar{o}_{j}, \bar{u}_{j}]$ represents a tuple of descriptions with each corresponding to a span---a subsequence of tokens---in $I$. Accordingly, $\bar{a}_{1:m}$
represents the description tuple sequence that we refer to as $\bar{a}$ for brevity. 
We also define $\bar{A}$ as all possible description tuple sequences of $I$, thus $\bar{a}\in{\bar{A}}$.

\begin{equation}
    \label{eq:phr2act}
    p(a_{j}|s_{j},t_{1:n})=\sum_{\bar{A}}p(a_{j}|\bar{a},s_{j},t_{1:n})p(\bar{a}|s_{j},t_{1:n})
\end{equation}

Because $a_j$ is independent of the rest of the instruction given its current screen $s_j$ and description $\bar{a}_j$, and $\bar{a}$ is only related to the instruction $t_{1:n}$, we can simplify (\ref{eq:phr2act}) as (\ref{eq:phr2act2}).

\begin{equation}
    \label{eq:phr2act2}
    p(a_{j}|s_{j},t_{1:n})=\sum_{\bar{A}}p(a_{j}|\bar{a}_j,s_{j})p(\bar{a}|t_{1:n})
\end{equation}

We define $\hat{a}$ as the most likely description of actions for $t_{1:n}$.

\begin{equation}
    \label{eq:best_desc}
    \begin{aligned}
    \hat{a} &=\argmax_{\bar{a}} p(\bar{a}|t_{1:n}) \\
            &=\argmax_{\bar{a}_{1:m}} \prod_{j=1}^{m}p(\bar{a}_{j}|\bar{a}_{<j},t_{1:n})
    \end{aligned}
\end{equation}

\noindent
This defines the action phrase-extraction model, which is then used by the grounding model:

\begin{equation}
    \label{eq:best_desc2}
    p(a_{j}|s_{j},t_{1:n}) \approx p(a_{j}|\hat{a}_j,s_{j})p(\hat{a}_j|\hat{a}_{<j},t_{1:n})
\end{equation}

\begin{equation}
    \label{eq:ground2}
    p(a_{1:m}|t_{1:n},S) \approx \prod_{j=1}^{m}p(a_{j}|\hat{a}_j,s_{j})p(\hat{a}_j|\hat{a}_{<j},t_{1:n})
\end{equation}

\noindent
$p(\hat{a}_j|\hat{a}_{<j},t_{1:n})$ identifies the description tuples for each action. $p(a_{j}|\hat{a}_j,s_{j})$ grounds each description to an executable action given the screen.

\section{Data}
\label{sec:dataset}

The ideal dataset would have natural instructions that have been executed by people using the UI.
Such data can be collected by having annotators perform tasks according to instructions on a mobile platform, but this is difficult to scale. It requires significant investment to instrument: different versions of apps have different presentation and behaviors, and apps must be installed and configured for each task. Due to this, we create a small dataset of this form, \pixelhelp, for full task evaluation. For model training at scale, we create two other datasets: \howto\ for action phrase extraction and \ricosynth\ for grounding. Our datasets are targeted for English. We hope that starting with a high-resource language will pave the way to creating similar capabilities for other languages.

\subsection{\pixelhelp\ Dataset}
\label{sec:test_data}

\begin{figure}
  \includegraphics[width=\columnwidth]{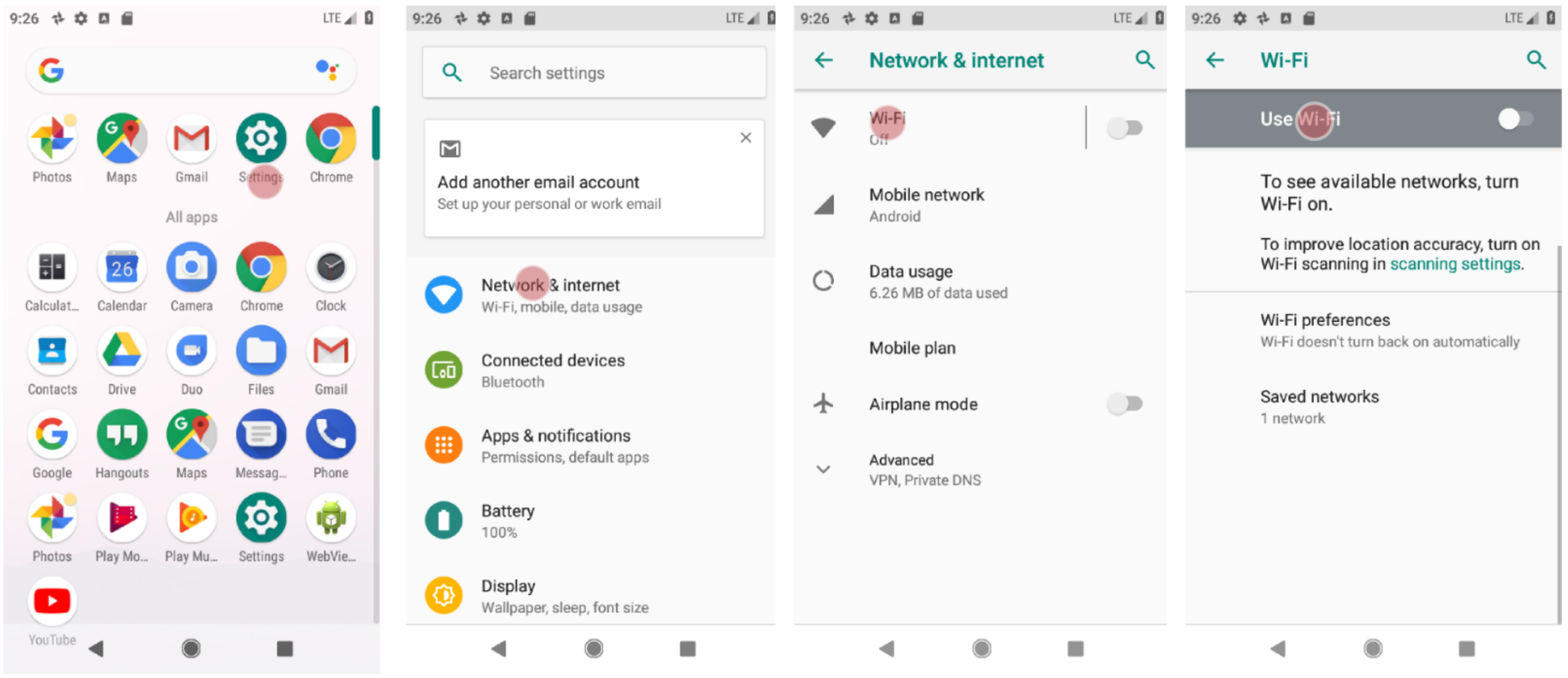}
  \caption{\pixelhelp\ example: \textit{Open your device's Settings app. Tap Network \& internet. Click Wi-Fi. Turn on Wi-Fi.}. The instruction is paired with actions, each of which is shown as a red dot on a specific screen.}
  \label{fig:how_to_example}
\end{figure}

Pixel Phone Help pages\footnote{\url{https://support.google.com/pixelphone}} provide instructions for performing common tasks on Google Pixel phones such as \textit{switch Wi-Fi settings} (Fig. \ref{fig:how_to_example}) or \textit{check emails}. Help pages can contain multiple tasks, with each task consisting of a sequence of steps. We pulled instructions from the help pages and kept ones that can be automatically executed. Instructions that requires additional user input such as \textit{Tap the app you want to uninstall} are discarded. Also, instructions that involve actions on a physical button such as \textit{Press the Power button for a few seconds} are excluded because these events cannot be executed on mobile platform emulators.

We instrumented a logging mechanism on a Pixel Phone emulator and had human annotators perform each task on the emulator by following the full instruction. The logger records every user action, including the type of touch events that are triggered, each object being manipulated, and screen information such as view hierarchies. Each item thus includes the instruction input, $t_{1:n}$, the screen for each step of task, $s_{1:m}$, and the target action performed on each screen, $a_{1:m}$. 

In total, \pixelhelp\ includes 187 multi-step instructions of 4 task categories: 88 general tasks, such as configuring accounts, 38 Gmail tasks, 31 Chrome tasks, and 30 Photos related tasks. The number of steps ranges from two to eight, with a median of four. Because it has both natural instructions and grounded actions, we reserve \pixelhelp\ for evaluating full task performance.

\subsection{\howto\ Dataset}
\label{sec:natural_data}

No datasets exist that support learning the action phrase extraction model, $p(\hat{a}_j|\hat{a}_{<j},t_{1:n})$, for mobile UIs. To address this, we extracted English instructions for operating Android devices by processing web pages to identify candidate instructions for how-to questions such as \textit{how to change the input method for Android}. A web crawling service scrapes instruction-like content from various websites. We then filter the web contents using both heuristics and manual screening by annotators.

Annotators identified phrases in each instruction that describe executable actions. They were given a tutorial on the task and were instructed to skip instructions that are difficult to understand or label. For each component in an action description, they select the span of words that describes the component using a web annotation interface (details are provided in the appendix). The interface records the start and end positions of each marked span. Each instruction was labeled by three annotators: three annotators agreed on 31\% of full instructions and at least two agreed on 84\%. For the consistency at the tuple level, the agreement across all the annotators is 83.6\% for operation phrases, 72.07\% for object phrases, and 83.43\% for input phrases. The discrepancies are usually small, e.g., a description marked as \textit{your Gmail address} or \textit{Gmail address}.

The final dataset includes 32,436 data points from 9,893 unique How-To instructions and split into training (8K), validation (1K) and test (900). All test examples have perfect agreement across all three annotators for the \textit{entire} sequence.  In total, there are 190K operation spans, 172K object spans, and 321 input spans labeled. The lengths of the instructions range from 19 to 85 tokens, with median of 59. They describe a sequence of actions from one to 19 steps, with a median of 5.

\subsection{\ricosynth\ Dataset}
\label{sec:synthetic_data}

Training the grounding model, $p(a_{j}|\hat{a}_j,s_{j})$  involves pairing action tuples $a_j$ along screens $s_j$ with action description $\hat{a}_{j}$. It is very difficult to collect such data at scale. To get past the bottleneck, we exploit two properties of the task to generate a \textit{synthetic} command-action dataset, \ricosynth. First, we have precise structured and visual knowledge of the UI layout, so we can spatially relate UI elements to each other and the overall screen. Second, a  grammar grounded in the UI can cover many of the commands and kinds of reference needed for the problem. This does not capture all manners of interacting conversationally with a UI, but it proves effective for training the grounding model.

Rico is a public UI corpus with 72K Android UI screens mined from 9.7K Android apps \citep{Deka:2017:Rico}. Each screen in Rico comes with a screenshot image and a view hierarchy of a collection of UI objects. Each individual object, $c_{j,k}$, has a set of properties, including its name (often an English phrase such as \textit{Send}), type (e.g., \texttt{Button}, \texttt{Image} or \texttt{Checkbox}), and bounding box position on the screen.  We manually removed screens whose view hierarchies do not match their screenshots by asking annotators to visually verify whether the bounding boxes of view hierarchy leaves match each UI object on the corresponding screenshot image. This filtering results in 25K unique screens.

For each screen, we randomly select UI elements as target objects and synthesize commands for operating them. We generate multiple commands to capture different expressions describing the operation $\hat{r}_j$ and the target object $\hat{o}_j$. For example, the \texttt{Tap} operation can be referred to as \textit{tap}, \textit{click}, or \textit{press}. The template for referring to a target object has slots \texttt{Name}, \texttt{Type}, and \texttt{Location}, which are instantiated  using the following strategies:

\begin{itemize}[nosep]
    \item \textit{Name-Type}: the target's name and/or type (\textit{the OK button} or \textit{OK}).
    \item \textit{Absolute-Location}: the target's screen location (\textit{the menu at the top right corner}).
    \item \textit{Relative-Location}: the target's relative location to other objects (\textit{the icon to the right of Send}).
\end{itemize}

\noindent
Because all commands are synthesized, the span that describes each part of an action, $\hat{a}_j$ with respect to $t_{1:n}$, is known. Meanwhile, ${a}_j$ and $s_j$, the actual action and the associated screen, are present because the constituents of the action are synthesized. In total, \ricosynth\ contains 295,476 single-step synthetic commands for operating 177,962 different target objects across 25,677 Android screens.

\begin{figure*}[t]
  \centering
  \includegraphics[width=.9\textwidth]{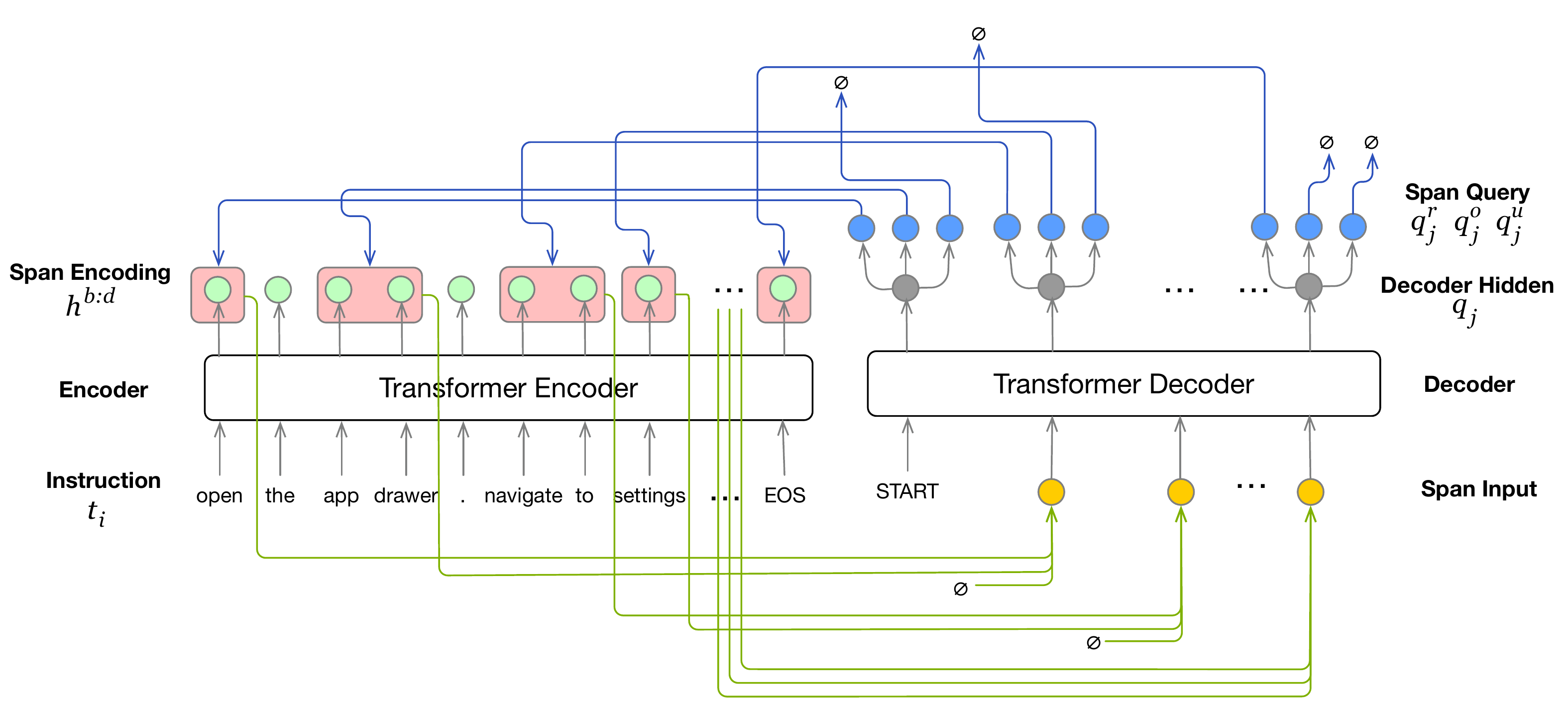}
  \caption{The Phrase Tuple Extraction model encodes the instruction's token sequence and then outputs a tuple sequence by querying into all possible spans of the encoded sequence. Each tuple contains the span positions of three phrases in the instruction that describe the action's operation, object and optional arguments, respectively, at each step. $\varnothing$ indicates the phrase is missing in the instruction and is represented by a special span encoding.}
  \label{fig:phrase}
\end{figure*}

\section{Model Architectures}
\label{sec:architectures}

Equation \ref{eq:ground2} has two parts. $p(\hat{a}_j|\hat{a}_{<j},t_{1:n})$ finds the best phrase tuple that describes the action at the $j$th step given the instruction token sequence. $p(a_{j}|\hat{a}_j,s_{j})$ computes the probability of an executable action $a_j$ given the best description of the action, $\hat{a}_j$, and the screen $s_j$ for the $j$th step. 

\subsection{Phrase Tuple Extraction Model}
\label{sec:phrase}

A common choice for modeling the conditional probability $p(\bar{a}_j|\bar{a}_{<j},t_{1:n})$ (see Equation \ref{eq:best_desc}) are encoder-decoders such as
LSTMs \citep{Hochreiter:1997:LSM:1246443.1246450} and Transformers \citep{,DBLP:journals/corr/VaswaniSPUJGKP17}. The output of our model corresponds to positions in the input sequence, so our architecture is closely related to Pointer Networks \citep{NIPS2015_5866}. 

Figure \ref{fig:phrase} depicts our model. An encoder $g$ computes a latent representation $h_{1:n}{\in}{R^{n\times{|h|}}}$ 
of the tokens from their embeddings: $h_{1:n}{=}g(e(t_{1:n}))$.
A decoder $f$ then generates the hidden state $q_{j}{=}f(q_{<j},\bar{a}_{<j},h_{1:n})$ which is used to compute a query vector that locates each phrase of a tuple ($\bar{r}_{j}$, $\bar{o}_{j}$, $\bar{u}_{j}$) at each step.
%
%
$\bar{a}_j{=}[\bar{r}_j, \bar{o}_j, \bar{u}_j]$ and they are assumed conditionally independent given previously extracted phrase tuples and the instruction, so $p(\bar{a}_j|\bar{a}_{<j},t_{1:n}){=}\prod_{\bar{y}\in{\{\bar{r},\bar{o},\bar{u}\}}}p(\bar{y}_j|\bar{a}_{<j},t_{1:n})$. 

Note that $\bar{y}_j\in{\{\bar{r}_j,\bar{o}_j,\bar{u}_j\}}$ denotes a specific span for $y\in\{r, o, u\}$ in the action tuple at step $j$. We therefore rewrite $\bar{y}_j$ as $y_{j}^{b:d}$ to explicitly indicate that it corresponds to the span for $r$, $o$ or $u$, starting at the $b$th position and ending at the $d$th position in the instruction, $1{\leq}{b}{<}d{\leq}{n}$. We now parameterize the conditional probability as:

\begin{equation}
    \label{eq:locate}
    \begin{split}
    &p(y_j^{b:d}|\bar{a}_{<j},t_{1:n}) = \mbox{softmax}(\alpha(q_{j}^{y}, h^{b:d})) \\
    &y\in\{r, o, u\}
    \end{split}
\end{equation}
\noindent
As shown in Figure \ref{fig:phrase}, $q_{j}^{y}$ indicates task-specific query vectors for $y{\in}\{r, o, u\}$. They are computed as $q_{j}^{y}{=}\phi(q_{j}, \theta_{y})W_{y}$, a multi-layer perceptron followed by a linear transformation. $\theta_{y}$ and $W_{y}$ are trainable parameters. We use separate parameters for each of $r$, $o$ and $u$. $W_{y}\in{R^{|\phi_y|\times{|h|}}}$ where $|\phi_y|$ is the output dimension of the multi-layer perceptron. 
The alignment function $\alpha(\cdot)$ scores how a query vector $q_{j}^{y}$ matches a span whose vector representation $h^{b:d}$ is computed from encodings $h_{b:d}$.

\textbf{Span Representation.}
There are a quadratic number of possible spans given a token sequence \citep{lee-etal-2017-end}, so it is important to design a fixed-length representation $h^{b:d}$ of a variable-length token span  that can be \textit{quickly} computed. Beginning-Inside-Outside (BIO) \citep{ramshaw-marcus-1995-text}--commonly used to indicate spans in tasks such as named entity recognition--marks whether each token is beginning, inside, or outside a span. However, BIO is not ideal for our task because subsequences for describing different actions can overlap, e.g., in \textit{click X and Y}, \textit{click} participates in both actions \textit{click X} and \textit{click Y}. In our experiments we consider several recent, more flexible span representations \citep{DBLP:journals/corr/LeeKP016, lee-etal-2017-end,pmlr-v97-li19e} and show their impact in Section \ref{sec:training}.

With fixed-length span representations, we can use common alignment techniques in neural networks \citep{Bahdanau2014NeuralMT,luong-etal-2015-effective}. We use the dot product between the query vector and the span representation: $\alpha(q_{j}^{y}, h^{b:d}){=}q_{j}^{y} \cdot h^{b:d}$
At each step of decoding, we feed the previously decoded phrase tuples, $\bar{a}_{<j}$ into the decoder. 
We can use the concatenation of the vector representations of the three elements in a phrase tuple or the sum their vector representations as the input for each decoding step. 
The entire phrase tuple extraction model is trained by minimizing the softmax cross entropy loss between the predicted and ground-truth spans of a sequence of phrase tuples. 

\begin{figure*}[t]
  \centering
  \includegraphics[width=.95\textwidth]{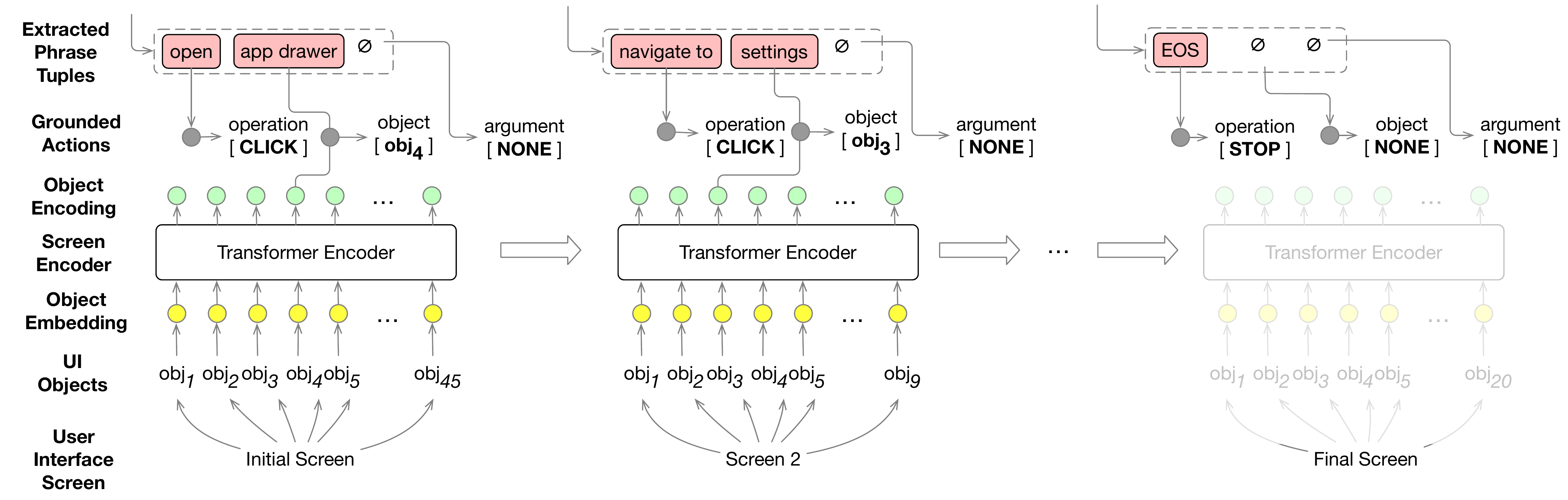}
  \caption{The Grounding model grounds each phrase tuple extracted by the Phrase Extraction model as an operation type, a screen-specific object ID, and an argument if present, based on a contextual representation of UI objects for the given screen. A grounded action tuple can be automatically executed.}
  \label{fig:ground}
\end{figure*}

\subsection{Grounding Model}
\label{sec:grounding}

Having computed the sequence of tuples that best describe each action, we connect them to executable actions based on the screen at each step with our grounding model (Fig. \ref{fig:ground}).
In step-by-step instructions, each part of an action is often clearly stated. Thus, we assume the probabilities of the operation $r_j$, object $o_j$, and argument $u_j$ are independent given their description and the screen. 

\begin{equation}
    \label{eq:ground3}
    \begin{split}
     p(a_{j}|\hat{a}_j,s_{j}) &= p([r_{j},o_{j},u_{j}]|[\hat{r}_j,\hat{o}_j,\hat{u}_j],s_{j})\\ &=p(r_{j}|\hat{r}_j,s_j)p(o_{j}|\hat{o}_j,s_j)p(u_{j}|\hat{u}_j,s_j)\\ &=p(r_{j}|\hat{r}_j)p(o_{j}|\hat{o}_j,s_j)
    \end{split}
\end{equation}

\noindent
We simplify with two assumptions: (1) an operation is often fully described by its instruction without relying on the screen information and (2) in mobile interaction tasks, an argument is only present for the \texttt{Text} operation, so $u_j{=}\hat{u}_j$. We parameterize $p(r_{j}|\hat{r}_j)$ as a feedforward neural network:

\begin{equation}
    \label{eq:operation}
    p(r_j|\hat{r}_j)=\mbox{softmax}(\phi(\hat{r}_{j}^{'}, \theta_{r})W_{r})
\end{equation}

\noindent
$\phi(\cdot)$ is a multi-layer perceptron with trainable parameters $\theta_r$. $W^r{\in}{R^{|\phi_r|\times{|r|}}}$ is also trainable, where $|\phi_r|$ is the output dimension of the $\phi(\cdot, \theta_r)$ and $|r|$ is the vocabulary size of the operations. $\phi(\cdot)$ takes the sum of the embedding vectors of each token in the operation description $\hat{r}_j$ as the input: $\hat{r}_{j}^{'}{=}\sum_{k=b}^{d}e(t_{k})$ where $b$ and $d$ are the start and end positions of $\hat{r}_j$ in the instruction.

Determining $o_{j}$ is to select a UI object from a variable-number of objects on the screen, $c_{j,k}\in{s_j}$ where $1{\leq}{k}{\leq}{|s_j|}$, based on the given object description, $\hat{o}_j$. We parameterize the conditional probability as a deep neural network with a softmax output layer taking logits from an alignment function:

\begin{equation}
    \label{eq:expand_ground}
    \begin{split}
    p(o_{j}|\hat{o}_{j},s_{j}) &=p(o_j=c_{j,k}|\hat{o}_{j},c_{j,1:|s_j|},\lambda_j) \\       &=\mbox{softmax}(\alpha(\hat{o}_{j}^{'}, c_{j,k}^{'}))
    \end{split}
\end{equation}

The alignment function $\alpha(\cdot)$ scores how the object description vector $\hat{o}_j^{'}$ matches the latent representation of each UI object, $c_{j,k}^{'}$. This can be as simple as the dot product of the two vectors. The latent representation $\hat{o}_j^{'}$ is acquired with a multi-layer perceptron followed by a linear projection: 

\begin{equation}
\label{eq:o_j}
    \hat{o}_{j}^{'}=\phi(\sum_{k=b}^{d}e(t_{k}),\theta_{o})W_{o}
\end{equation}

\noindent
$b$ and $d$ are the start and end index of the object description $\hat{o}_j$. $\theta_{o}$ and $W_{o}$ are trainable parameters with $W_{o}{\in}{R^{|\phi_{o}|\times{|o|}}}$, where $|\phi_{o}|$ is the output dimension of $\phi(\cdot,\theta_{o})$ and $|o|$ is the dimension of the latent representation of the object description.

\textbf{Contextual Representation of UI Objects.}
To compute latent representations of each candidate object, $c_{j,k}^{'}$, we use both the object's properties and its context, i.e., the structural relationship with other objects on the screen. There are different ways for encoding a variable-sized collection of items that are structurally related to each other, including Graph Convolutional Networks (GCN) \citep{pmlr-v48-niepert16} and Transformers \citep{DBLP:journals/corr/VaswaniSPUJGKP17}. GCNs use an adjacency matrix predetermined by the UI structure to regulate how the latent representation of an object should be affected by its neighbors. Transformers allow each object to carry its own positional encoding, and the relationship between objects can be learned instead. 

The input to the Transformer encoder is a combination of the \textit{content embedding} and the \textit{positional encoding} of each object. The content properties of an object include its name and type. We compute the content embedding of by concatenating the name embedding, which is the average embedding of the bag of tokens in the object name, and the type embedding. The positional properties of an object include both its spatial position and structural position. The spatial positions include the top, left, right and bottom screen coordinates of the object. We treat each of these coordinates as a discrete value and represent it via an embedding. Such a feature representation for coordinates was used in ImageTransformer to represent pixel positions in an image \citep{pmlr-v80-parmar18a}. The spatial embedding of the object is the sum of these four coordinate embeddings. To encode structural information, we use the index positions of the object in the preorder and the postorder traversal of the view hierarchy tree, and represent these index positions as embeddings in a similar way as representing coordinates. The content embedding is then summed with positional encodings to form the embedding of each object. We then feed these object embeddings into a Transformer encoder model to compute the latent representation of each object, $c_{j,k}^{'}$. 

The grounding model is trained by minimizing the cross entropy loss between the predicted and ground-truth object and the loss between the predicted and ground-truth operation.

\section{Experiments}
\label{sec:experiments}

Our goal is to develop models and datasets to map multi-step instructions into automatically executable actions given the screen information. As such, we use \pixelhelp's paired natural instructions and action-screen sequences solely for testing. In addition, we investigate the model quality on phrase tuple extraction tasks, which is a crucial building block for the overall grounding quality\footnote{Our model code is released at \url{https://github.com/google-research/google-research/tree/master/seq2act}.}.

\subsection{Datasets and Metrics}
\label{sec:datasets}

We use two metrics that measure how a predicted tuple sequence matches the ground-truth sequence.

\begin{itemize}[nosep]
    \item \textit{Complete Match}: The score is $1$ if two sequences have the same length and have the identical tuple $[\hat{r}_j,\hat{o}_j,\hat{u}_j]$ at each step, otherwise $0$.
    \item \textit{Partial Match}: The number of steps of the predicted sequence that match the ground-truth sequence divided by the length of the ground-truth sequence (ranging between $0$ and $1$).
\end{itemize}

We train and validate using \howto\ and \ricosynth, and evaluate on \pixelhelp. During training, single-step synthetic command-action examples are dynamically stitched to form sequence examples with a certain length distribution. To evaluate the full task, we use Complete and Partial Match on grounded action sequences $a_{1:m}$ where $a_j{=}[r_j, o_j, u_j]$.

The token vocabulary size is 59K, which is compiled from both the instruction corpus and the UI name corpus. There are 15 UI types, including 14 common UI object types, and a type to catch all less common ones. The output vocabulary for operations include \texttt{CLICK}, \texttt{TEXT}, \texttt{SWIPE} and \texttt{EOS}.

\subsection{Model Configurations and Results}
\label{sec:training}

\textbf{Tuple Extraction.} For the action-tuple extraction task, we use a 6-layer Transformer for both the encoder and the decoder. We evaluate three different span representations. Area Attention \citep{pmlr-v97-li19e} provides a parameter-free representation of each possible span (one-dimensional area), by summing up the encoding of each token in the subsequence: $h^{b:d}=\sum_{k=b}^{d}h_{k}$. The representation of each span can be computed in constant time invariant to the length of the span, using a summed area table. Previous work concatenated the encoding of the start and end tokens as the span representation, $h^{b:d}=[h_b; h_d]$ \citep{DBLP:journals/corr/LeeKP016} and a generalized version of it \citep{lee-etal-2017-end}. We evaluated these three options and implemented the representation in \newcite{lee-etal-2017-end} using a summed area table similar to the approach in area attention for fast computation. For hyperparameter tuning and training details, refer to the appendix. 

\begin{table}
\centering
\begin{tabular}{l|cc}
\textbf{Span Rep. $h^{b:d}$} & \textbf{Partial} & \textbf{Complete} \\ 
\hline
Sum Pooling $\sum_{k=b}^{d}h_{k}$ & 92.80 & 85.56 \\
StartEnd Concat$[h_{b}; h_{d}]$ & 91.94 & 84.56 \\
$[h_{b}; h_{d}, \hat{e}^{b:d}, \phi(d-b)]$ & 91.11 & 84.33 \\
\end{tabular}
\caption{\label{tab:span} \howto\ phrase tuple extraction test results using different span representations $h^{b:d}$ in (\ref{eq:locate}). $\hat{e}^{b:d}{=}\sum_{k=b}^{d}w(h_{k})e(t_k)$, where $w(\cdot)$ is a learned weight function for each token embedding \cite{lee-etal-2017-end}. See the pseudocode for fast computation of these in the appendix.}
\vspace{-10pt}
\end{table}

Table \ref{tab:span} gives results on \howto's test set. All the span representations perform well. Encodings of each token from a Transformer already capture sufficient information about the entire sequence, so even only using the start and end encodings yields strong results. Nonetheless, area attention provides a small boost over the others. As a new dataset, there is also considerable headroom remaining, particularly for complete match.

\textbf{Grounding.} For the grounding task, we compare Transformer-based screen encoder for generating object representations $h^{b:d}$ with two baseline methods based on graph convolutional networks. The \textit{Heuristic} baseline matches extracted phrases against object names directly using BLEU scores. \textit{Filter-1 GCN} performs graph convolution without using adjacent nodes (objects), so the representation of each object is computed only based on its own properties. \textit{Distance GCN} uses the distance between objects in the view hierarchy, i.e., the number of edges to traverse from one object to another following the tree structure. This contrasts with the traditional GCN definition based on adjacency, but is needed because UI objects are often leaves in the tree; as such, they are not adjacent to each other structurally but instead are connected through non-terminal (container) nodes. Both Filter-1 GCN and Distance GCN use the same number of parameters (see the appendix for details).

To train the grounding model, we first train the Tuple Extraction sub-model on \howto\ and \ricosynth. For the latter, only language related features (commands and tuple positions in the command) are used in this stage, so screen and action features are not involved. We then freeze the Tuple Extraction sub-model and train the grounding sub-model on \ricosynth\ using both the command and screen-action related features. The screen token embeddings of the grounding sub-model share weights with the Tuple Extraction sub-model. 

\begin{table}
\centering
\begin{tabular}{l|cc}
\textbf{Screen Encoder} & \textbf{Partial} & \textbf{Complete} \\ 
\hline
Heuristic & 62.44 & 42.25 \\
Filter-1 GCN & 76.44 & 52.41 \\
Distance GCN & 82.50 & 59.36 \\
Transformer & 89.21 & 70.59 \\
\end{tabular}
\caption{\label{tab:srn} \pixelhelp\ grounding accuracy. The differences are statistically significant based on t-test over 5 runs ($p<0.05$).}
\end{table}

Table \ref{tab:srn} gives full task performance on \pixelhelp. The Transformer screen encoder achieves the best result with 70.59\% accuracy on Complete Match and 89.21\% on Partial Match, which sets a strong baseline result for this new dataset while leaving considerable headroom. The GCN-based methods perform poorly, which shows the importance of contextual encodings of the information from other UI objects on the screen. Distance GCN does attempt to capture context for UI objects that are structurally close; however, we suspect that the distance information that is derived from the view hierarchy tree is noisy because UI developers can construct the structure differently for the same UI.\footnote{While it is possible to directly use screen visual data for grounding, detecting UI objects from raw pixels is nontrivial. It would be ideal to use both structural and visual data.} As a result, the strong bias introduced by the structure distance does not always help. Nevertheless, these models still outperformed the heuristic baseline that achieved 62.44\% for partial match and 42.25\% for complete match.

\subsection{Analysis}
\label{sec:analysis}

\begin{figure}
  \includegraphics[width=\columnwidth]{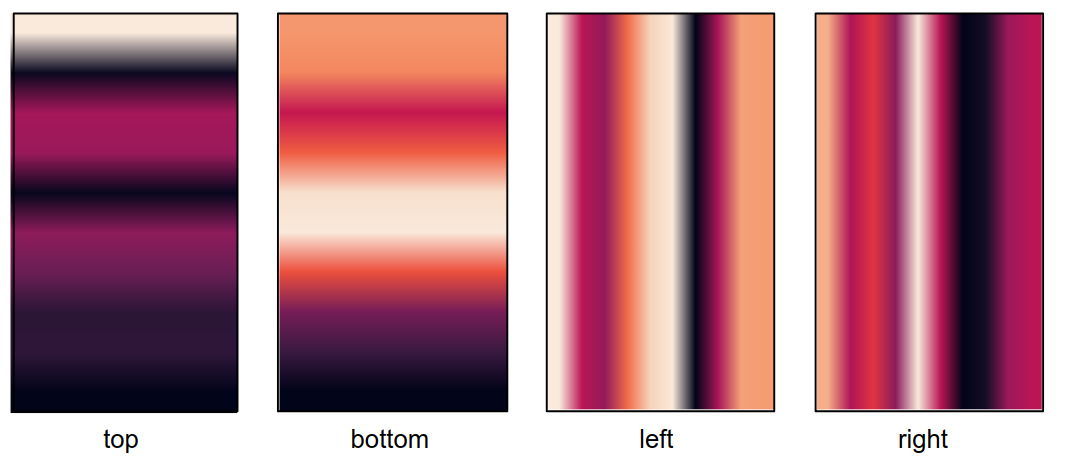}
  \caption{Correlation between location-related words in instructions and object screen position embedding.}
  \label{fig:location_grd}
\end{figure}

To explore how the model grounds an instruction on a screen, we analyze the relationship between words in the instruction language that refer to specific locations on the screen, and actual positions on the UI screen. We first extract the embedding weights from the trained phrase extraction model for words such as \textit{top}, \textit{bottom}, \textit{left} and \textit{right}. These words occur in object descriptions such as \textit{the check box at the top of the screen}. We also extract the embedding weights of object screen positions, which are used to create object positional encoding. We then calculate the correlation between word embedding and screen position embedding using cosine similarity. Figure \ref{fig:location_grd} visualizes the correlation as a heatmap, where brighter colors indicate higher correlation. The word \textit{top} is strongly correlated with the top of the screen, but the trend for other location words is less clear. While \textit{left} is strongly correlated with the left side of the screen, other regions on the screen also show high correlation. This is likely because \textit{left} and \textit{right} are not only used for referring to absolute locations on the screen, but also for relative spatial relationships, such as \textit{the icon to the left of the button}.
For \textit{bottom}, the strongest correlation does not occur at the very bottom of the screen because many UI objects in our dataset do not fall in that region. The region is often reserved for system actions and the on-screen keyboard, which are not covered in our dataset.

The phrase extraction model passes phrase tuples to the grounding model. When phrase extraction is incorrect, it can be difficult for the grounding model to predict a correct action. One way to mitigate such cascading errors is using the hidden state of the phrase decoding model at each step, $q_j$. Intuitively, $q_j$ is computed with the access to the encoding of each token in the instruction via the Transformer encoder-decoder attention, which can potentially be a more robust span representation. However, in our early exploration, we found that grounding with $q_j$ performs stunningly well for grounding \ricosynth\ validation examples, but performs poorly on \pixelhelp. The learned hidden state likely captures characteristics in the synthetic instructions and action sequences that do not manifest in \pixelhelp.  As such, using the hidden state to ground remains a challenge when learning from unpaired instruction-action data.

The phrase model failed to extract correct steps for 14 tasks in \pixelhelp. In particular, it resulted in extra steps for 11 tasks and extracted incorrect steps for 3 tasks, but did not skip steps for any tasks. These errors could be caused by different language styles manifested by the three datasets. Synthesized commands in \ricosynth\space tend to be brief. Instructions in \howto\space seem to give more contextual description and involve diverse language styles, while \pixelhelp\space often has a more consistent language style and gives concise description for each step.

\section{Related Work}
Previous work \citep{Branavan:2009:RLM:1687878.1687892,branavan-etal-2010-reading,liu2018workflow,gur2018learning} investigated approaches for grounding natural language on desktop or web interfaces. \newcite{manuvinakurike-etal-2018-edit} contributed a dataset for mapping natural language instructions to actionable image editing commands in Adobe Photoshop. Our work focuses on a new domain of grounding natural language instructions into executable actions on mobile user interfaces. This requires addressing modeling challenges due to the lack of paired natural language and action data, which we supply by harvesting rich instruction data from the web and synthesizing UI commands based on a large scale Android corpus.

Our work is related to semantic parsing, particularly efforts for generating executable outputs such as SQL queries \cite{suhr-etal-2018-learning}. It is also broadly related to language grounding in the human-robot interaction literature where human dialog results in robot actions \cite{Khayrallah2015NaturalLF}.

Our task setting is closely related to work on language-conditioned navigation, where an agent executes an instruction as a sequence of movements \citep{Chen:2011:LIN:2900423.2900560,Mei:2016:LAW:3016100.3016289,Misra:17instructions,Anderson:2018:VLN,Chen19:touchdown}. Operating user interfaces is similar to navigating the physical world in many ways. A mobile platform consists of millions of apps that each is implemented by different developers independently. Though platforms such as Android strive to achieve interoperability (e.g., using Intent or AIDL mechanisms), apps are more often than not built by convention and do not expose programmatic ways for communication. As such, each app is opaque to the outside world and the only way to manipulate it is through its GUIs. These hurdles while working with a vast array of existing apps are like physical obstacles that cannot be ignored and must be negotiated contextually in their given environment.

\section{Conclusion}
\label{sec:conclusion}
Our work provides an important first step on the challenging problem of grounding natural language instructions to mobile UI actions. Our decomposition of the problem means that progress on either can improve full task performance. For example, action span extraction is related to both semantic role labeling \citep{he-etal-2018-jointly} and extraction of multiple facts from text \cite{jiang-etal-2019-multi} and could benefit from innovations in span identification and multitask learning. Reinforcement learning that has been applied in previous grounding work may help improve out-of-sample prediction for grounding in UIs and improve direct grounding from hidden state representations. Lastly, our work provides a technical foundation for investigating user experiences in language-based human computer interaction.

\section*{Acknowledgements}
We would like to thank our anonymous reviewers for their insightful comments that improved the paper. Many thanks to the Google Data Compute team, especially Ashwin Kakarla and Muqthar Mohammad for their help with the annotations, and Song Wang, Justin Cui and Christina Ou for their help on early data preprocessing.

\bibliography{anthology,acl2020}

\begin{thebibliography}{28}
\expandafter\ifx\csname natexlab\endcsname\relax\def\natexlab#1{#1}\fi

\bibitem[{Anderson et~al.(2018)Anderson, Wu, Teney, Bruce, Johnson,
  S{\"u}nderhauf, Reid, Gould, and van~den Hengel}]{Anderson:2018:VLN}
Peter Anderson, Qi~Wu, Damien Teney, Jake Bruce, Mark Johnson, Niko
  S{\"u}nderhauf, Ian Reid, Stephen Gould, and Anton van~den Hengel. 2018.
\newblock Vision-and-language navigation: Interpreting visually-grounded
  navigation instructions in real environments.
\newblock In \emph{Proceedings of the IEEE Conference on Computer Vision and
  Pattern Recognition (CVPR)}.

\bibitem[{Bahdanau et~al.(2014)Bahdanau, Cho, and
  Bengio}]{Bahdanau2014NeuralMT}
Dzmitry Bahdanau, Kyunghyun Cho, and Yoshua Bengio. 2014.
\newblock Neural machine translation by jointly learning to align and
  translate.
\newblock \emph{CoRR}, abs/1409.0473.

\bibitem[{Branavan et~al.(2009)Branavan, Chen, Zettlemoyer, and
  Barzilay}]{Branavan:2009:RLM:1687878.1687892}
S.~R.~K. Branavan, Harr Chen, Luke~S. Zettlemoyer, and Regina Barzilay. 2009.
\newblock \href {http://dl.acm.org/citation.cfm?id=1687878.1687892}
  {Reinforcement learning for mapping instructions to actions}.
\newblock In \emph{Proceedings of the Joint Conference of the 47th Annual
  Meeting of the ACL and the 4th International Joint Conference on Natural
  Language Processing of the AFNLP: Volume 1 - Volume 1}, ACL '09, pages
  82--90, Stroudsburg, PA, USA. Association for Computational Linguistics.

\bibitem[{Branavan et~al.(2010)Branavan, Zettlemoyer, and
  Barzilay}]{branavan-etal-2010-reading}
S.R.K. Branavan, Luke Zettlemoyer, and Regina Barzilay. 2010.
\newblock \href {https://www.aclweb.org/anthology/P10-1129} {Reading between
  the lines: Learning to map high-level instructions to commands}.
\newblock In \emph{Proceedings of the 48th Annual Meeting of the Association
  for Computational Linguistics}, pages 1268--1277, Uppsala, Sweden.
  Association for Computational Linguistics.

\bibitem[{Chen and Mooney(2011)}]{Chen:2011:LIN:2900423.2900560}
David~L. Chen and Raymond~J. Mooney. 2011.
\newblock \href {http://dl.acm.org/citation.cfm?id=2900423.2900560} {Learning
  to interpret natural language navigation instructions from observations}.
\newblock In \emph{Proceedings of the Twenty-Fifth AAAI Conference on
  Artificial Intelligence}, AAAI'11, pages 859--865. AAAI Press.

\bibitem[{Chen et~al.(2019)Chen, Suhr, Misra, and Artzi}]{Chen19:touchdown}
Howard Chen, Alane Suhr, Dipendra Misra, and Yoav Artzi. 2019.
\newblock Touchdown: Natural language navigation and spatial reasoning in
  visual street environments.
\newblock In \emph{Conference on Computer Vision and Pattern Recognition}.

\bibitem[{Deka et~al.(2017)Deka, Huang, Franzen, Hibschman, Afergan, Li,
  Nichols, and Kumar}]{Deka:2017:Rico}
Biplab Deka, Zifeng Huang, Chad Franzen, Joshua Hibschman, Daniel Afergan, Yang
  Li, Jeffrey Nichols, and Ranjitha Kumar. 2017.
\newblock Rico: A mobile app dataset for building data-driven design
  applications.
\newblock In \emph{Proceedings of the 30th Annual Symposium on User Interface
  Software and Technology}, UIST '17.

\bibitem[{Gur et~al.(2019)Gur, Rueckert, Faust, and
  Hakkani-Tur}]{gur2018learning}
Izzeddin Gur, Ulrich Rueckert, Aleksandra Faust, and Dilek Hakkani-Tur. 2019.
\newblock \href {https://openreview.net/forum?id=BJemQ209FQ} {Learning to
  navigate the web}.
\newblock In \emph{International Conference on Learning Representations}.

\bibitem[{He et~al.(2018)He, Lee, Levy, and Zettlemoyer}]{he-etal-2018-jointly}
Luheng He, Kenton Lee, Omer Levy, and Luke Zettlemoyer. 2018.
\newblock \href {https://doi.org/10.18653/v1/P18-2058} {Jointly predicting
  predicates and arguments in neural semantic role labeling}.
\newblock In \emph{Proceedings of the 56th Annual Meeting of the Association
  for Computational Linguistics (Volume 2: Short Papers)}, pages 364--369,
  Melbourne, Australia. Association for Computational Linguistics.

\bibitem[{Hochreiter and
  Schmidhuber(1997)}]{Hochreiter:1997:LSM:1246443.1246450}
Sepp Hochreiter and J\"{u}rgen Schmidhuber. 1997.
\newblock \href {https://doi.org/10.1162/neco.1997.9.8.1735} {Long short-term
  memory}.
\newblock \emph{Neural Comput.}, 9(8):1735--1780.

\bibitem[{Jiang et~al.(2019)Jiang, Zhao, Qin, Liu, Chawla, and
  Jiang}]{jiang-etal-2019-multi}
Tianwen Jiang, Tong Zhao, Bing Qin, Ting Liu, Nitesh Chawla, and Meng Jiang.
  2019.
\newblock \href {https://doi.org/10.18653/v1/D19-1029} {Multi-input
  multi-output sequence labeling for joint extraction of fact and condition
  tuples from scientific text}.
\newblock In \emph{Proceedings of the 2019 Conference on Empirical Methods in
  Natural Language Processing and the 9th International Joint Conference on
  Natural Language Processing (EMNLP-IJCNLP)}, pages 302--312, Hong Kong,
  China. Association for Computational Linguistics.

\bibitem[{Khayrallah et~al.(2015)Khayrallah, Trott, and
  Feldman}]{Khayrallah2015NaturalLF}
Huda Khayrallah, Sean Trott, and Jerome Feldman. 2015.
\newblock Natural language for human robot interaction.

\bibitem[{Lee et~al.(2017)Lee, He, Lewis, and Zettlemoyer}]{lee-etal-2017-end}
Kenton Lee, Luheng He, Mike Lewis, and Luke Zettlemoyer. 2017.
\newblock \href {https://doi.org/10.18653/v1/D17-1018} {End-to-end neural
  coreference resolution}.
\newblock In \emph{Proceedings of the 2017 Conference on Empirical Methods in
  Natural Language Processing}, pages 188--197, Copenhagen, Denmark.
  Association for Computational Linguistics.

\bibitem[{Lee et~al.(2016)Lee, Kwiatkowski, Parikh, and
  Das}]{DBLP:journals/corr/LeeKP016}
Kenton Lee, Tom Kwiatkowski, Ankur~P. Parikh, and Dipanjan Das. 2016.
\newblock \href {http://arxiv.org/abs/1611.01436} {Learning recurrent span
  representations for extractive question answering}.
\newblock \emph{CoRR}, abs/1611.01436.

\bibitem[{Li et~al.(2019)Li, Kaiser, Bengio, and Si}]{pmlr-v97-li19e}
Yang Li, Lukasz Kaiser, Samy Bengio, and Si~Si. 2019.
\newblock \href {http://proceedings.mlr.press/v97/li19e.html} {Area attention}.
\newblock In \emph{Proceedings of the 36th International Conference on Machine
  Learning}, volume~97 of \emph{Proceedings of Machine Learning Research},
  pages 3846--3855, Long Beach, California, USA. PMLR.

\bibitem[{Liu et~al.(2018)Liu, Guu, Pasupat, Shi, and Liang}]{liu2018workflow}
E.~Z. Liu, K.~Guu, P.~Pasupat, T.~Shi, and P.~Liang. 2018.
\newblock Reinforcement learning on web interfaces using workflow-guided
  exploration.
\newblock In \emph{International Conference on Learning Representations
  (ICLR)}.

\bibitem[{Luong et~al.(2015)Luong, Pham, and
  Manning}]{luong-etal-2015-effective}
Thang Luong, Hieu Pham, and Christopher~D. Manning. 2015.
\newblock \href {https://doi.org/10.18653/v1/D15-1166} {Effective approaches to
  attention-based neural machine translation}.
\newblock In \emph{Proceedings of the 2015 Conference on Empirical Methods in
  Natural Language Processing}, pages 1412--1421, Lisbon, Portugal. Association
  for Computational Linguistics.

\bibitem[{Manuvinakurike et~al.(2018)Manuvinakurike, Brixey, Bui, Chang, Kim,
  Artstein, and Georgila}]{manuvinakurike-etal-2018-edit}
Ramesh Manuvinakurike, Jacqueline Brixey, Trung Bui, Walter Chang, Doo~Soon
  Kim, Ron Artstein, and Kallirroi Georgila. 2018.
\newblock \href {https://www.aclweb.org/anthology/L18-1683} {Edit me: A corpus
  and a framework for understanding natural language image editing}.
\newblock In \emph{Proceedings of the Eleventh International Conference on
  Language Resources and Evaluation ({LREC}-2018)}, Miyazaki, Japan. European
  Languages Resources Association (ELRA).

\bibitem[{Mei et~al.(2016)Mei, Bansal, and
  Walter}]{Mei:2016:LAW:3016100.3016289}
Hongyuan Mei, Mohit Bansal, and Matthew~R. Walter. 2016.
\newblock \href {http://dl.acm.org/citation.cfm?id=3016100.3016289} {Listen,
  attend, and walk: Neural mapping of navigational instructions to action
  sequences}.
\newblock In \emph{Proceedings of the Thirtieth AAAI Conference on Artificial
  Intelligence}, AAAI'16, pages 2772--2778. AAAI Press.

\bibitem[{Misra et~al.(2017)Misra, Langford, and Artzi}]{Misra:17instructions}
Dipendra Misra, John Langford, and Yoav Artzi. 2017.
\newblock \href {https://doi.org/10.18653/v1/D17-1106} {Mapping instructions
  and visual observations to actions with reinforcement learning}.
\newblock In \emph{Proceedings of the Conference on Empirical Methods in
  Natural Language Processing}, pages 1004--1015.

\bibitem[{Niepert et~al.(2016)Niepert, Ahmed, and Kutzkov}]{pmlr-v48-niepert16}
Mathias Niepert, Mohamed Ahmed, and Konstantin Kutzkov. 2016.
\newblock \href {http://proceedings.mlr.press/v48/niepert16.html} {Learning
  convolutional neural networks for graphs}.
\newblock In \emph{Proceedings of The 33rd International Conference on Machine
  Learning}, volume~48 of \emph{Proceedings of Machine Learning Research},
  pages 2014--2023, New York, New York, USA. PMLR.

\bibitem[{Parmar et~al.(2018)Parmar, Vaswani, Uszkoreit, Kaiser, Shazeer, Ku,
  and Tran}]{pmlr-v80-parmar18a}
Niki Parmar, Ashish Vaswani, Jakob Uszkoreit, Lukasz Kaiser, Noam Shazeer,
  Alexander Ku, and Dustin Tran. 2018.
\newblock \href {http://proceedings.mlr.press/v80/parmar18a.html} {Image
  transformer}.
\newblock In \emph{Proceedings of the 35th International Conference on Machine
  Learning}, volume~80 of \emph{Proceedings of Machine Learning Research},
  pages 4055--4064, Stockholmsmässan, Stockholm Sweden. PMLR.

\bibitem[{Ramshaw and Marcus(1995)}]{ramshaw-marcus-1995-text}
Lance Ramshaw and Mitch Marcus. 1995.
\newblock \href {https://www.aclweb.org/anthology/W95-0107} {Text chunking
  using transformation-based learning}.
\newblock In \emph{Third Workshop on Very Large Corpora}.

\bibitem[{Sarsenbayeva(2018)}]{sarsenbayeva:2018}
Zhanna Sarsenbayeva. 2018.
\newblock Situational impairments during mobile interaction.
\newblock In \emph{Proceedings of the ACM on Interactive, Mobile, Wearable and
  Ubiquitous Technologies}, pages 498--503.

\bibitem[{Suhr et~al.(2018)Suhr, Iyer, and Artzi}]{suhr-etal-2018-learning}
Alane Suhr, Srinivasan Iyer, and Yoav Artzi. 2018.
\newblock \href {https://doi.org/10.18653/v1/N18-1203} {Learning to map
  context-dependent sentences to executable formal queries}.
\newblock In \emph{Proceedings of the 2018 Conference of the North {A}merican
  Chapter of the Association for Computational Linguistics: Human Language
  Technologies, Volume 1 (Long Papers)}, pages 2238--2249, New Orleans,
  Louisiana. Association for Computational Linguistics.

\bibitem[{Szeliski(2010)}]{Szeliski:2010:CVA:1941882}
Richard Szeliski. 2010.
\newblock \emph{Computer Vision: Algorithms and Applications}, 1st edition.
\newblock Springer-Verlag, Berlin, Heidelberg.

\bibitem[{Vaswani et~al.(2017)Vaswani, Shazeer, Parmar, Uszkoreit, Jones,
  Gomez, Kaiser, and Polosukhin}]{DBLP:journals/corr/VaswaniSPUJGKP17}
Ashish Vaswani, Noam Shazeer, Niki Parmar, Jakob Uszkoreit, Llion Jones,
  Aidan~N. Gomez, Lukasz Kaiser, and Illia Polosukhin. 2017.
\newblock \href {http://arxiv.org/abs/1706.03762} {Attention is all you need}.
\newblock \emph{CoRR}, abs/1706.03762.

\bibitem[{Vinyals et~al.(2015)Vinyals, Fortunato, and Jaitly}]{NIPS2015_5866}
Oriol Vinyals, Meire Fortunato, and Navdeep Jaitly. 2015.
\newblock \href {http://papers.nips.cc/paper/5866-pointer-networks.pdf}
  {Pointer networks}.
\newblock In C.~Cortes, N.~D. Lawrence, D.~D. Lee, M.~Sugiyama, and R.~Garnett,
  editors, \emph{Advances in Neural Information Processing Systems 28}, pages
  2692--2700. Curran Associates, Inc.

\end{thebibliography}
\bibliographystyle{acl_natbib}

\begin{figure*}
  \centering
  \includegraphics[width=\textwidth]{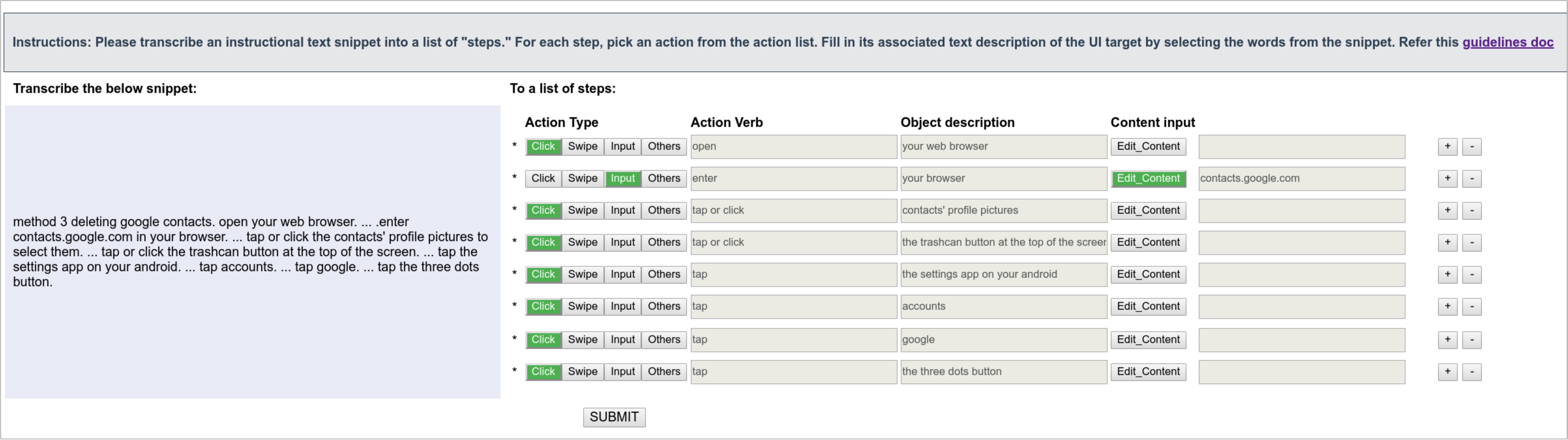}
  \caption{The web interface for annotators to label action phrase spans in an \howto\ instruction.}
  \label{fig:plugin}
\end{figure*}
\appendix
\section{Data}
We present the additional details and analysis of the datasets. To label action phrase spans for the \howto\ dataset, 21 annotators (9 males and 12 females, 23 to 28 years old) were employed as contractors. They were paid hourly wages that are competitive for their locale. They have standard rights as contractors. They were native English speakers, and rated themselves 4 out of 5 regarding their familiarity with Android (1: not familiar and 5: very familiar). 

Each annotator is presented a web interface, depicted in Figure \ref{fig:plugin}. The instruction to be labeled is shown on the left of the interface. From the instruction, the annotator is asked to extract a sequence of action phrase tuples on the right, providing one tuple per row. Before a labeling session, an annotator is asked to go through the annotation guidelines, which are also accessible throughout the session.

To label each tuple, the annotator first indicates the type of operation (Action Type) the step is about by selecting from \texttt{Click}, \texttt{Swipe}, \texttt{Input} and \texttt{Others} (the catch-all category). The annotator then uses a mouse to select the phrase in the instruction for ``Action Verb'' (i.e., operation description) and for ``object description''. A selected phrase span is automatically shown in the corresponding box and the span positions in the instruction are recorded. If the step involves an additional argument, the annotator clicks on ``Content Input'' and then marks a phrase span in the instruction (see the second row). Once finished with creating a tuple, the annotator moves onto the next tuple by clicking the ``+'' button on the far right of the interface along the row, which inserts an empty tuple after the row. The annotator can delete a tuple (row) by clicking the ``-'' button on the row. Finally, the annotator clicks on the ``Submit'' button at the bottom of the screen to finish a session. 

The lengths of the instructions range from 19 to 85 tokens, with median of 59, 
and they describe a sequence of actions from 1 to 19 steps, with a median of 5. 
Although the description for operations tend to be short (most of them are one to two words), the description for objects can vary dramatically in length, ranging from 1 to 19. 
The large range of description span lengths requires an efficient algorithm to compute its representation.




\section{Computing Span Representations}

We evaluated three types of span representations. Here we give details on how each representation is computed. For sum pooling, we use the implementation of area attention \citep{pmlr-v97-li19e} that allows constant time computation of the representation of each span by using summed area tables. The TensorFlow implementation of the representation is available on Github\footnote{\url{https://github.com/tensorflow/tensor2tensor/blob/master/tensor2tensor/layers/area_attention.py}}.

\begin{algorithm}
\caption{Compute the Start-End Concat span representation for all spans in parallel.}
 \label{alg:start_end}
 \KwIn{A tensor $H$ in shape of $[L, D]$ that represents a sequence of vector with length $L$ and depth $D$.}
 \KwOut{representation of each span, $U$.}

{\bf{Hyperparameter}}: max span width $M$.

{\bf Init} start \& end tensor: $S\leftarrow H$, $E\leftarrow H$\;
\For{$m=1,\cdots, M - 1$}{
   $S^{'} \leftarrow H[:-m,:]$ \;
   $E^{'} \leftarrow H[m:,:]$ \;
   $S\leftarrow [S\ \ S^{'}]$, concat on the 1st dim\;
   $E\leftarrow [E\ \ E^{'}]$, concat on the 1st dim\;
  }
  $U\leftarrow [S\ \ E]$, concat on the last dim\;
 \KwRet $U$.
\end{algorithm}

Algorithm \ref{alg:start_end} gives the recipe for Start-End Concat \citep{DBLP:journals/corr/LeeKP016} using Tensor operations. The advanced form \cite{lee-etal-2017-end} takes two other features: the weighted sum over all the token embedding vectors within each span and a span length feature. The span length feature is trivial to compute in a constant time. However, computing the weighted sum of each span can be time consuming if not carefully designed. We decompose the computation as a set of summation-based operations (see Algorithm \ref{alg:weighted_sum} and \ref{alg:span_sum}) so as to use summed area tables \citep{Szeliski:2010:CVA:1941882}, which was been used in \newcite{pmlr-v97-li19e} for constant time computation of span representations. These pseudocode definitions are designed based on Tensor operations, which are highly optimized and fast.

\begin{algorithm}
\caption{Compute the weighted embedding sum of each span in parallel, using ComputeSpanVectorSum defined in Algorithm \ref{alg:span_sum}.}
 \label{alg:weighted_sum}
 \KwIn{Tensors $H$ and $E$ are the hidden and embedding vectors of a sequence of tokens respectively, in shape of $[L,D]$ with length $L$ and depth $D$.}
 \KwOut{weighted embedding sum, $\hat{X}$.}

{\bf{Hyperparameter}}: max span length $M$.

Compute token weights $A$: $A \leftarrow \mbox{exp}(\phi(H, \theta)W)$ where $\phi(\cdot)$ is a multi-layer perceptron with trainable parameters $\theta$, followed by a linear transformation $W$. $A\in{R^{L\times{1}}}$\;

$E^{'} \leftarrow E \otimes A$ where $\otimes$ is element-wise multiplication. The last dim of $A$ is broadcast\;

$\hat{E} \leftarrow \mbox{ComputeSpanVectorSum}(E^{'})$\;

$\hat{A} \leftarrow \mbox{ComputeSpanVectorSum}(A)$\;

$\hat{X} \leftarrow \hat{E} \oslash \hat{A}$ where $\oslash$ is element-wise division. The last dim of $\hat{A}$ is broadcast\;

\KwRet $\hat{X}$.

\end{algorithm}

\begin{algorithm}
\caption{ComputeSpanVectorSum.}
 \label{alg:span_sum}
 \KwIn{A tensor $G$ in shape of $[L, D]$.}
 \KwOut{Sum of vectors of each span, $U$.}

{\bf{Hyperparameter}}: max span length $M$.

{\bf Compute} integral image $I$ by cumulative sum along the first dimension over $G$\; 
$I \leftarrow [\mbox{0}\ \ I]$, padding zero to the left\;
\For{$m=0,\cdots, M-1$}{
   $I_1 \leftarrow I[m+1:,:]$ \;
   $I_2 \leftarrow I[:-m-1,:]$ \;
   $\bar{I} \leftarrow I_1-I_2$ \;
   $U\leftarrow [U\ \ \bar{I}]$, concat on the first dim\;
 }
 \KwRet $U$.
\end{algorithm}

\section{Details for Distance GCN}

Given the structural distance between two objects, based on the view hierarchy tree, we compute the strength of how these objects should affect each other by applying a Gaussian kernel to the distance, as shown the following (Equation \ref{eq:gcn}).

\begin{equation}
\label{eq:gcn}
\mbox{Adjacency}(o_i,o_j)=\frac{1}{\sqrt{2\pi\sigma^{2}}}\mbox{exp}(-\frac{d(o_i,o_j)^2}{2\sigma^2})
\end{equation}

\noindent
where $d(o_i, o_j)$ is the distance between object $o_i$ and $o_j$, and $\sigma$ is a constant. With this definition of soft adjacency, the rest of the computation follows the typical GCN \citep{pmlr-v48-niepert16}.

\section{Hyperparameters \& Training}
We tuned all the models on a number of hyperparameters, including the token embedding depth, the hidden size and the number of hidden layers, the learning rate and schedule, and the dropout ratios. 
We ended up using 128 for the embedding and hidden size for all the models. Adding more dimensions does not seem to improve accuracy and slows down training. 

For the phrase tuple extraction task, we used 6 hidden layers for Transformer encoder and decoder, with 8-head self and encoder-decoder attention, for all the model configurations. We used 10\% dropout ratio for attention, layer preprocessing and relu dropout in Transformer. We followed the learning rate schedule detailed previously \citep{DBLP:journals/corr/VaswaniSPUJGKP17}, with an increasing learning rate to 0.001 for the first 8K steps followed by an exponential decay. All the models were trained for 1 million steps with a batch size of 128 on a single Tesla V100 GPU, which took 28 to 30 hours.

For the grounding task, Filter-1 GCN and Distance GCN used 6 hidden layers with ReLU for nonlinear activation and 10\% dropout ratio at each layer. Both GCN models use a  smaller peak learning rate of 0.0003. The Transformer screen encoder also uses 6 hidden layers but uses a much larger dropout ratio: ReLU dropout of 30\%, attention dropout of 40\%, and layer preprocessing dropout of 20\%, with a peak learning rate of 0.001. All the grounding models were trained for 250K steps on the same hardware.

\end{document}